\documentclass[conference]{IEEEtran}
\IEEEoverridecommandlockouts
\usepackage{cite}
\usepackage{amsmath,amssymb,amsfonts}
\usepackage{graphicx}
\usepackage{textcomp}
\usepackage{xcolor}

\usepackage{mathtools, nccmath}
\usepackage{algorithm,algpseudocode}

\usepackage{xcolor, soul}
\sethlcolor{lightgray}
\usepackage{subfig}

\usepackage{multirow}
\newcommand{\niton}{\not\owns} 

\def\BibTeX{{\rm B\kern-.05em{\sc i\kern-.025em b}\kern-.08em
    T\kern-.1667em\lower.7ex\hbox{E}\kern-.125emX}}
\begin{document}

\title{Real-Time Long Range Trajectory Replanning for MAVs in the Presence of Dynamic Obstacles\\
}

\makeatletter
\newcommand{\linebreakand}{%
  \end{@IEEEauthorhalign}
  \hfill\mbox{}\par
  \mbox{}\hfill\begin{@IEEEauthorhalign}
}
\makeatother
\author{
  \IEEEauthorblockN{Geesara Kulathunga, Roman Fedorenko, Sergey Kopylov, Alexandr Klimchik}
  \IEEEauthorblockA{
   \textit{Center for Technologies in Robotics and Mechatronics Components, Innopolis University, Russia} \\
    ggeesara@gmail.com, r.fedorenko@innopolis.ru, s.kopylov@innopolis.ru, a.klimchik@innopolis.ru}
 
%
}

\maketitle

\begin{abstract}
Real-time long-range local planning is a challenging task, especially in the presence of dynamics obstacles. We propose a complete system which is capable of performing the local replanning in real-time. Desired trajectory is needed in the system initialization phase; system starts initializing sub-components of the system including point cloud processor, trajectory estimator and planner. Afterwards, the multi-rotary aerial vehicle starts moving on the given trajectory. When it detects obstacles, it replans the trajectory from the current pose to pre-defined distance incorporating the desired trajectory. Point cloud processor is employed to identify the closest obstacles around the vehicle. For replanning, Rapidly-exploring Random Trees (RRT*) is used with two modifications which allow planning the trajectory in milliseconds scales. Once we replanned the desired path, velocity components(x,y and z) and yaw rate are calculated. Those values are sent to the controller at a constant frequency to maneuver the vehicle autonomously. Finally, we have evaluated each of the components separately and tested the complete system in the simulated and real environments.
\end{abstract}


\section{Introduction}

\begin{figure}[!ht]
   \subfloat[Experimenting proposed solution in the filed]{ \includegraphics[width=0.9\linewidth, height=3.2cm]{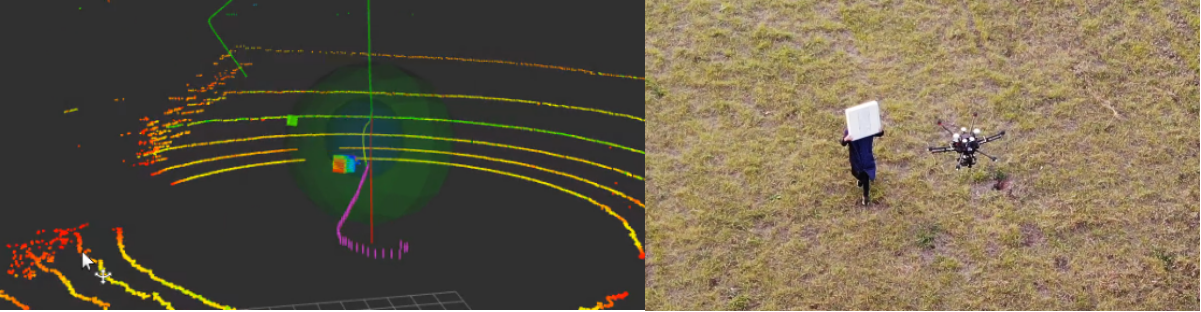} \label{f:separation_on_bfs}}\\
   \subfloat[Experimenting on our simulator. Green color ellipsoid depicts the search space whereas blue color sphere depicts the second search space which is used to generate random point when MAV is close to obstacles. Light green color line and red line denote target trajectory and trajectory to be completed respectively. Replanned trajectory is shown in yellow colored line]{ \includegraphics[width=0.9\linewidth, height=3.2cm]{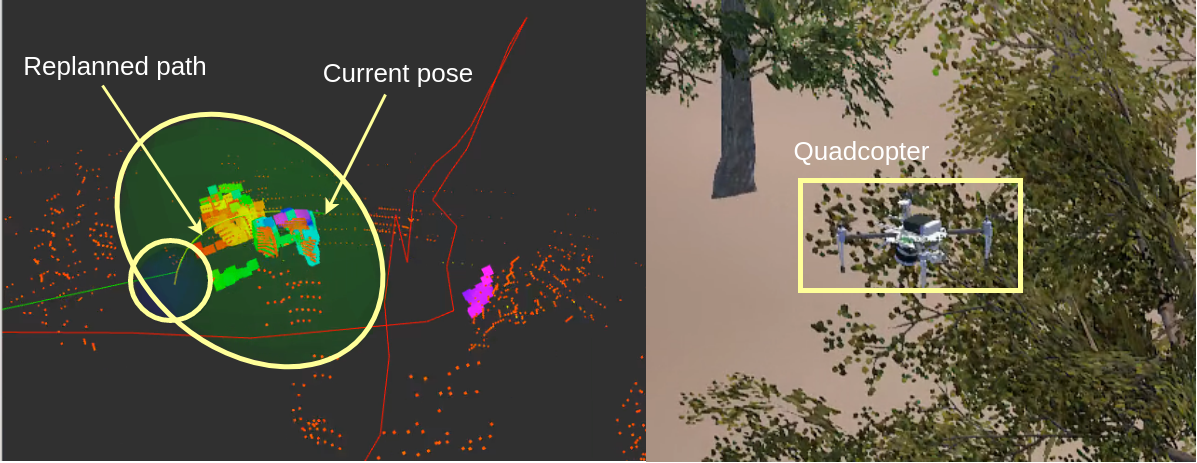} \label{f:projected_on_camera}}
  \caption[]{Experimental result on our simulator and field }
  \label{f:main_steps}
\end{figure}

Multirotor Aerial Vehicles (MAVs) draw attention not only among the researches but also from the general audience due to the emerging trustworthiness towards the real-world applicability in the recent past.
Rescue operations in dangerous places where humans are not able to reach, surveillance and delivering of goods are a few good examples of MAVs applications. In all the scenarios, robust planning is the key component to obtain a promising result in spite of the other constraints (e.g., state estimating, controlling, etc). On certain occasions, MAV may fly into completely unknown zone due to lack of understanding the environment. It could be a dangerous where MAV cloud move into tree canopy in which obstacles are not captured within sensors' field of view (FoV). Hence, this should be avoided at all cost. Thus, along with global planning, task can not be completed. Applying local replanning considering dynamic obstacles is the correct way to approach this problem.  

At the present time, most of the commercial MAVs do not have robust local replanning capabilities. Even it supports, capacities are at the primitive level. Local replanning while avoiding obstacles is a must to have for safe navigation, in the sense that understanding of the environment is essential to have proper reasoning about obstacles. In the recent works, Octomap is used for building a profoundly accurate map of the environment. On the contrary, high computational power and considerable memory resources are required to incrementally update the Octomap. Moreover, extending the area of the map is quite challenging while keeping high level of accuracy because of high demanding of computing resources.       

In the proposed solution, a set of consecutive point clouds is incrementally concatenated instead of building a persistent map. Afterwards, the point cloud is extracted within a pre-defined radius of the sphere relative to the current position of the MAV. To identify obstacles within the extracted point could, we used the method that is proposed in \cite{prathap2019ground}. Then, the instant obstacles map is constructed as an Rtree~\cite{guttman1984r} which utilizes for planning. When obstacles are encountered inside pre-defined zone, path segment within this zone is reprojected by replanner while considering the initial target trajectory. For planning, we modified the original RRT* algorithm introducing quadratic search space and incorporating the target trajectory by which MAV is to manoeuvre.

\textbf{Our Contributions}
\begin{enumerate}
    \item Modified original RRT* algorithm introducing quadratic search space. Moreover, the random sample selection process of RRT* is improved by incorporating the target trajectory. These changes helped to gain real-time performance in a cluttered environment.
    \item Proposing a software framework that executes in parallel to achieve real-time performance and further reduce the execution time by using code-level optimization
\end{enumerate}

\section{Related Work}

In this section, we discuss robotics motion planning
literature with focus on methods of local replanning. We split our discussion into path planning, trajectory smoothing and detecting of regions of interest followed by environment representation from point cloud.

\textbf{Robotic Motion Planning} is about using prior information to generate a set of control inputs to maneuver the robot from its initial position to the goal position complemented by having full awareness of the environmental conditions (i.e., appropriate reactions when robot sees the obstacles). This is achieved in two stages: globally and locally. In global motion planning trajectory is generated based on static obstacles. On the contrary, local motion planning is dynamic planning in which trajectory is constructed based on current sensor information about the environment. 

\textbf{Path Planning} has two branches: graph and sampling based. A* and Dijkstra's are the most commonly used algorithms in graph-based search. A grid (discretization of continuous space) is constructed as the search space. Then, this grid is considered as a graph which is utilized for finding a path. On the other hand, sampling-based algorithms including RRT* (Randomly exploring Random Tree)~\cite{karaman2011sampling}, PRMs (Probability Road Maps) and  EST (Expansive-Spaces Tree)~\cite{akinc2005probabilistic} work on continuous space and does not claim the optimality whereas graph-based algorithms do claim it. However, sampling-based techniques are computationally efficient which is the main reason those are suited for working with high-dimensional search spaces.

\textbf{Trajectory generation and smoothing} are closely related to each other. Trajectory generation can be classified into three subcategories: path planning followed by smoothing, optimization-based approaches (e.g., \cite{7759784,usenko2017real}) and motion primitive based approaches. Usually, smoothing is done using polynomial-spline rather than cubic spline. Cubic spline makes trajectory over smoother which gives low accuracy than a polynomial spline. Polynomial spline always keeps the original waypoints after smoothing. On the other hand, the main advantage  of motion primitive approaches such as (~\cite{pivtoraiko2013incremental,oleynikova2016continuous,barry2018high}) over the others is that its execution time varies on average in few milliseconds in a cluttered environment.

\textbf{Detecting regions of interest from point cloud and representing the  environment} are quite challenging tasks to perform in real-time. In this experiment, we are interested in extracting obstacles within a given sphere while neglecting all the noise information. Most of the previously proposed works, point cloud is fed into one of the following data structures to build a map of the environment: voxel grid ~\cite{elfes1989using}, octomap ~\cite{wurm2010octomap} and elevation map ~\cite{choi2012global}. However, all of the preceding methods are computationally expensive. On the other hand, F. Gao  \cite{7784290} utilizes globally-registered point cloud directly bypassing map building for local online planning where it first identifies the regions of interest from the point cloud and the construct instance map for further processing. Environment representation is mostly constructed as an occupancy map. Octomap~\cite{hornung2013octomap} which uses hierarchical octree is one of the ways of representing the search space. Octomap consists of small grids that are called 3D voxel~\cite{steinbrucker2014volumetric}. One of the benefits of octomap is that its voxels can be purged when they have same information. But it requires considerably high memory resources for storing. To overcome this memory limitation, authors of~\cite{niessner2013real} purposed voxel hashing which stores only the sign distance information. 

\textbf{Local Obstacle Avoidance} is the most crucial part of local planning. Reactive avoidance and map-based avoidance are the dominant methods. In reactive avoidance, planning relies on only current sensor information without building a persistence map of the surrounding. Map-based avoidance depends on a map which is constructed using sensor data or prior knowledge of the environment. Both of these can be fallen into local minima which essential to bypass. On the contrary, collision-free trajectory generation by building environment incrementally using octree is suggested in this work~\cite{chen2016online}. However, when the environment is cluttered, the preceding techniques do not work properly. Thus, in contrast to preceding methods, to mitigate those problems, Oleynikova et al.,~\cite{oleynikova2018safe} proposed a local exploration strategy based technique, to traverse in a cluttered environment. 

\section{Methodology}

Long-range trajectory planning while considering dynamic obstacles is always difficult owing to various reasons including error checking between current pose and desired trajectory, instantaneous reaction to avoid the dynamic obstacles and replan trajectory that satisfies real-time constraints. We propose a complete system that addresses all the preceding problems robustly. 

\subsection{System Architecture} 
\label{sec:sys_architecture}
\begin{figure}[ht]
\begin{center}
\includegraphics[width=.6\linewidth]{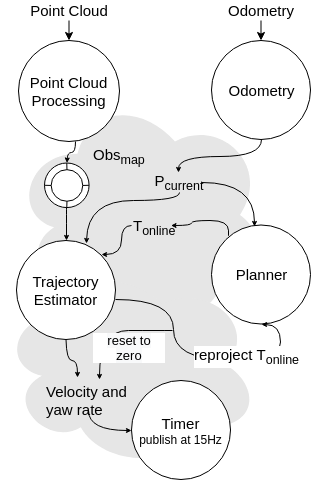}
\caption{\label{f:main_steps_of_trajectory_planner} High-level architecture of the system, complemented with all the shared variables which are denoted in bold font. Point cloud and odometry are taken as inputs whereas current velocity will be the output of the system, provided that planning thread is triggered when obstacles present close to MAV. Current velocity is reset to zero during the trajectory replanning}
\end{center}
\end{figure}

The system is designed as a multi-threaded application, in which five threads are utilized for achieving high-performance computation capability in real-time. The high-level view of the system is shown in Fig.~\ref{f:main_steps_of_trajectory_planner}. Point cloud processing thread processes raw point cloud from lidar and separate the regions that interest. Here, regions that interest indicate the surrounding obstacles. The output of point cloud processing thread is stored in a shared circular buffer ($Obs_{map}$), which size can be configured. The current pose of the MAV is one of the most important pieces of information to avoid obstacles in real-time and appropriately reproject the trajectory. Thus, the responsibility of the odometry thread is to update the current pose ($P_{current}$) of MAV at 15Hz; $P_{current}$ is also a shared variable. Trajectory estimator and Planner are interconnected with each other closely. The trajectory estimator ensures to reset velocity to zero when the obstacles present within a predefined radius ($obs_avoid_zone$), send a signal to the planner to reproject trajectory ($T_{projected}$) up to the $obs_avoid_zone$ and replace current trajectory ($T_{online}$) with $T_{projected}$. The controller expects a minimum number of control messages to operate properly. The 5th thread assures this requirement. It publishes current velocity and yaw rate at 15Hz. Details information can be found in following sections.  

\subsection{Trajectory Estimation}

\begin{figure*}[ht]
\begin{center}
\includegraphics[width=16cm]{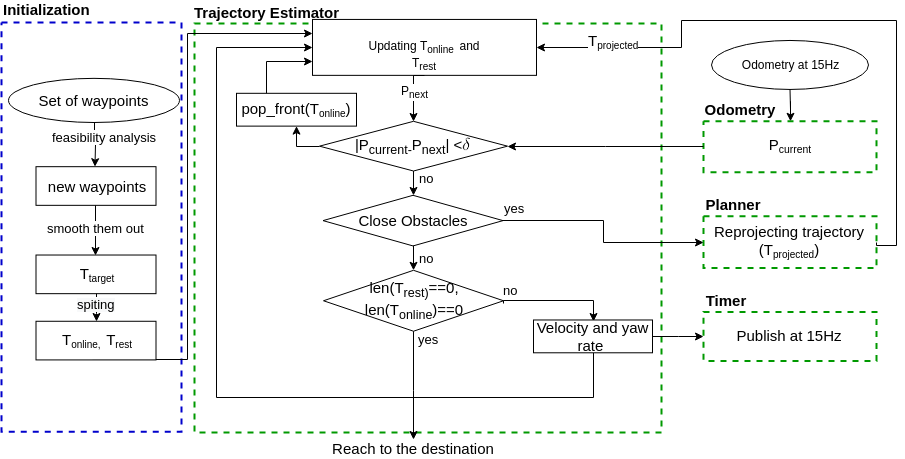}
\caption{\label{f:trajectory_planner} Trajectory estimation: trajectory ($T_{target}$) to be followed is constructed from provided a sequence of waypoints in the initialization stage; afterwards, $T_{online}$ and $T_{rest}$ are updated in parallel where $T_{online}$ referees to trajectory waypoints within a predefined distance ($replanning\_dis$) from the current pose ($P_{current}$) of MAV; Rest of the $T_{target}$ belongs to $T_{rest}$}
\end{center}
\end{figure*}

Initially, it is required to provide the trajectory in the form of waypoints as input to the system. Afterwards, a new set of waypoints is generated such that maximum distance between two waypoints is less or equals to $replanning\_dis$. Thus, the preceding process implies the feasibility of the target trajectory which defines as the feasibility analysis in.~\ref{f:trajectory_planner}. Lets define two successive waypoints $p_1$ and $p_2$ such that $|\mathbf{p_1}-\mathbf{p_2}| > obs\_avoid\_dis$. Hence, intermediate waypoints are generated in between them with coordinates according to Eq.~\ref{eq:spliting_waypoints}.

\begin{equation}
    \label{eq:spliting_waypoints}
     \begin{aligned}
        \mathbf{p\_inter}_{x} = obs\_avoid\_dis \cdot sin(\phi) \cdot cos(\theta) + \mathbf{p_1}_x \\
        \mathbf{p\_inter}_{y} = obs\_avoid\_dis \cdot sin(\phi) \cdot sin(\theta) + \mathbf{p_1}_y \\
        \mathbf{p\_inter}_{z} = obs\_avoid\_dis \cdot cos(\phi) + \mathbf{p_1}_z\\
        \theta = atan2(\mathbf{p}_y, \mathbf{p}_x), \quad \phi = atan2(\sqrt{\mathbf{p}_x^2 + \mathbf{p}_y^2}, \mathbf{p}_z) \\
         \mathbf{p} = \mathbf{p_2} - \mathbf{p_1}
    \end{aligned}   
\end{equation}

Afterwards, generated new waypoints are smoothed out to enhance the consistency of continuous trajectory ($T_{target}$). After experimenting on several approaches, normalized polynomial B-splines are chosen for trajectory smoothing. The normalized polynomial B-spline can be represented by the following Cox-de Boor recursive formula~\cite{de1972calculating,qin2000general}.

\begin{equation}
    \left\{\begin{matrix}
q_{j,3}(t) = \frac{t-t_j}{t_{j+2}-t_{j}} q_{j,2}(t) + \frac{t_{j+3}-t}{t_{j+3}-t_{j+1}} q_{j+1,2}(t)\\ 
q_{i,1}(t) = \left\{\begin{matrix}
1 & t \in [t_i, t_{i+1}) \\ 
 0& t \niton [t_i, t_{i+1})
\end{matrix}\right.
\end{matrix}\right.
\end{equation}

In general, this can be written in matrix form as follows:

\begin{equation}
[q_{i-2,3}(u) \; q_{i-1,3}(u) \; q_{i,3}(u)] = [1 \; u \; u^2]\textbf{M(i)} \begin{bmatrix}
p_{i-2}\\ 
p_{i-1}\\ 
p_{i}
\end{bmatrix}
\end{equation} where $u=\frac{t-t_{i}}{t_{i+1}-ti}, \; u \in [0,1)$, $p_j \mid j=\{i, i-1, i-2\}$ are consecutive control points (waypoints). According to the proof given in ~\cite{qin2000general}, \textbf{M(i)} stands for the $i$-th basis matrix of the B-Spline matrix which can be derived as follows:

\begin{equation*}
\begin{bmatrix}
\frac{({t_{i+1}}-t_i)}{(t_{i+1}-t_{i-1})} & \frac{t_i-t_{i-1}}{t_{i+1}-t_{i-1}} & 0 \\ 
\frac{-2(t_{i+1}-t_i)}{t_{i+1}-t_{i-1}} & \frac{2(t_{i+1}-t_i)}{t_{i+1}-t_{i-1}} & 0\\ 
\frac{t_{i+1}-t_i}{t_{i+1}-t_{i-1}} & -(t_{i+1}-t_i)(\frac{1}{t_{i+1}} + \frac{1}{t_{i+2}-t_i}) & m_{33} \\ 
\end{bmatrix}
\end{equation*} where $ m_{33} = \frac{t_{i+1}-t_i}{t_{i+2}- t_{i-1}}$.
Once $T_{target}$ is constructed, successive step is to split $T_{target}$ into $T_{online}$ and $T_{rest}$. $T_{online}$ is part of the trajectory within $replanning\_dis$ from  $T_{target}$'s start waypoint. Rest of $T_{target}$ belongs to $T_{rest}$. All the preceding steps up to this happens only once during the system initialization stage. Rest of the process is executed in parallel as shown in Fig.~\ref{f:main_steps_of_trajectory_planner}. Insight of trajectory estimator is given in Fig.~\ref{f:trajectory_planner}. Updating $T_{online}$ and $T_{rest}$ happens if and only if distance between first and last waypoint of $T_{online}$ is less than $replanning\_dis$ where distance is calculated summing up euclidean distances between consecutive waypoints in $T_{online}$. Updating of $T_{online}$ happens by taking consecutive waypoints from $T_{rest}$ such that total distance of $T_{online}$ is less than $replanning\_dis$ and fetch them to  $T_{online}$. 
This process happens as a sliding window fashion until the size of the both trajectories $T_{online}$ and $T_{rest}$ is not equal to zero. Thereby MAV comes to its final destination. 
$P_{next}$ denotes the first waypoint in $T_{online}$. Then if the L2 norm between $P_{current}$ and $P_{next}$ is less than $\delta$, remove $P_{next}$ from $T_{online}$ ($pop\_front(T_{online})$). 

In each iteration, neighbouring obstacles are scanned within a pre-defined radius. If obstacles are present, signal to the planner is sent. Then re-planning (projecting) of the trajectory starts from $P_{current}$ to end pose of $T_{online}$. Finally, we replace $T_{online}$ with $T_{projected}$. More descriptions of the planner's operation are given in the next section. 

\subsection{Improved RRT* Local Planner}
We have made two significant modifications to the original RRT* algorithm. These changes were made to reduce the execution time of the algorithm for finding a path.  

Let's define the problem statement adhere with~\cite{karaman2011sampling}. Search or configuration space is denoted with $X = (0, 1)^3$ where $X \in \mathbb{R}^3 $. Regions that are occupied by the obstacles are denoted with $X_{obs}$. 
Thus, $X \setminus X_{obs}$ becomes an open set where $cl(X\setminus X_{obs})$ can be defined as the traversable space ($X_{free}$). Here, $cl(.)$ is denotes the closure of set $X\setminus X_{obs}$. 
$X_{start}$ and $X_{goal}$ are the start and target position for planning where $X_{start} \in X_{free}$ and $X_{goal} \in X_{free}$. $\sigma: [0,1]^3 \rightarrow \mathbb{R} ^3$ denotes the waypoints of the planned path, provided that $\sigma (\tau) \in X_{free}$ for all $\tau \in [0, 1]$; This path is called as a obstacle free path. 
We are interested in finding a feasible path over an optimal path because the execution time of the planner is a critical factor where it should able to find a feasible path in real-time. 
Thus, feasible path exists if collision free path ($\sigma (\tau)$) is satisfies these two conditions: $\sigma (0) = X_{start}$ and $\sigma (1) \in cl(X_{goal})$. 
In order to define the cost of the feasible path, it is necessary to understand how RRT* algorithm works as given in Algorithm.~\ref{alg:rrt}, in which random sample $x_{rand} \sim U(X_{free})$ is generated in each iteration, where $U(.)$ stands for the uniform distribution. Afterwards, the parent vertex is selected by inspecting neighboring nodes (search within the pre-defined area) with respect to $x_{new}$. If new parent vertex can connect with current $x_{start}$, it becomes new parent. If $x_{new}$ is a vertex and it is connected to $x_{near}$, drop the existing edge and replace it with $x_{new}$. This process is called rewiring. 

Feasible path is denoted as $X_{start} \overset{f(x)}{\longrightarrow}$ $X_{goal}$, where $f(x)$ is the cost of the feasible path. Since RRT* is sampling-based, $f(x)$ depends on the traversable search space $X_{free}$. In the whole, $f(.)$ is a heuristic function which can be reformulated as follows: 
\begin{equation}
    f(x) = \hat{f}(x, X_{start}) + \hat{g}(x, X_{goal}), \quad x \in X_{free}
\end{equation}  where $\hat{f}$ and $\hat{g}$ both are admissible heuristic functions. Thus, $f(x)$ does not try to overestimate the cost of the path. When the search space ($X$) is bigger, it utilises a considerable amount of time to find a feasible path. We have noticed reducing the search space will help to converge the planner faster. Therefore, we introduce a new search space $X_{reduced}$ which is a subset of $X_{free}$. $X_{reduced}$ is a polate ellipsoidal search space as shown in Fig~\ref{fig:main_search_space} as similar to ~\cite{gammell2014informed}.

\begin{algorithm}
\caption{The main steps of RRT* algorithm. Colored lines with background highlighting reflect modifications of the original RRT*}
\label{alg:rrt}
\begin{algorithmic}[1]
\Procedure{RRT*}{}
 \State $V \gets {X_{start}}; E \gets \varnothing;$
 \State \hl{$\textcolor[rgb]{0.5,0,0}{Traj \gets Desired\_trajectory}$}
 \For {i=1,...,n}
    \State \hl{$\textcolor[rgb]{0.5,0,0}{x_{rand} \gets GetFreeSample_i}$}
    \State \hl{$\textcolor[rgb]{0.5,0,0}{x_{nearest} \gets Nearest(G=(V,E), x_{rand}, Traj);}$}
    \State $x_{new} \gets GetSteerPose(x_{nearest}, x_{rand});$
    \If{$ObstableFree(x_{nearest}, x_{new})$}
              \State $X_{near} \gets GetNearByVertices(x_{new})$
              \State $V \gets V \cup \{x_{near}\}$
              \State $ConnectShortestPath(x_{new}, x_{near})$
              \If {$IsEdge(x_{new})$}
                    \State $Rewine(x_{new}, x_{near})$
                    \State $E \gets E \cup \{(x_{new})\}$
              \EndIf
              \If {$CheckSolution()$}
                    \textbf{return} G = (V,E) 
              \EndIf
     \EndIf
 \EndFor
 \State \textbf{return} G = (V,E) 
\EndProcedure
\end{algorithmic}
\end{algorithm}

To prove $X_{reduced}$ search space has higher convergence rate over $X_{free}$, let's define the probability of selecting a random sample between $X_{free}$ and $X_{reduced}$ ($X_{reduced} \subset X_{free}$). 

\begin{figure}[ht]
\begin{center}
\includegraphics[width=0.65\linewidth]{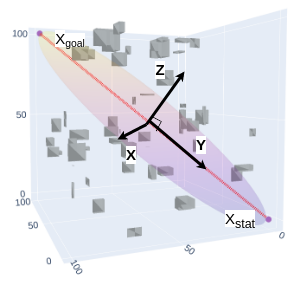}
\caption{The heuristic sampling space $X_{reduced}$ is depicted with ellipsoid in which transverse diameter is equal to $X_{goal}-X_{start}$ and conjugate diameter is a configurable parameter (i.e, 4m) on x and z direction}
\label{fig:main_search_space}
\end{center}
\end{figure}

\begin{equation}
     \begin{aligned}
    P(x \in X_{free}) \leqslant P(x \in X_{reduced})  = \frac{\lambda(X_{reduced})}{\lambda(X_{free})}\\
    \frac{\lambda(X_{reduced})}{\lambda(X_{free})} = \frac{3}{4} \pi d^2 \frac{|X_{goal}-X_{start}|}{\lambda(X_{free})}
    \end{aligned} 
    \label{eq:prababilty_prof}
\end{equation} 
where $\lambda(.)$ denotes the volume of the search space. As given in Eq.~\ref{eq:prababilty_prof}, $ \frac{\lambda(X_{reduced})}{\lambda(X_{free})}$ depicts the selecting a sample from $X_{reduced}$ always has higher probability. Transverse diameter and conjugate diameter of the ellipsoid are to be estimated when it is needed to replan the trajectory of MAV. Then, generating samples from $X_{reduced}$ is possible. 

Three-dimensional Gaussian bell can be represented as an ellipsoid where size and orientation of ellipsoid are described by the covariance matrix $\Sigma$. Thus, defining the $X_{reduced}$ can be seen as an analogy for defining three dimensional Gaussian bell, with R stands for rotation matrix and variances as a diagonal matrix $diag(\sigma_{xx}, \sigma_{yy}, \sigma_{zz})$. The relationship between $R,diag(\sigma_{xx}, \sigma_{yy}, \sigma_{zz})$ and $\Sigma$ is: 

\begin{equation}
    \Sigma = R\:\textbf{diag}(\sigma_{xx}, \sigma_{yy}, \sigma_{zz})\:R^t
\end{equation} where $\sigma_{xx} = |X_{goal} -X_{start}|$ and $\sigma_{yy} = \sigma_{zz} = d$. The rotation matrix R aligns $\mathbf{z}$ (0,0,1) to $\mathbf{X_{goal}} -\mathbf{X_{start}}$ and can be calculated as follows:
\begin{equation}
     \begin{aligned}
        R = I + [\mathbf{v}]_x + [\mathbf{v}]_x^2 \frac{(1-\mathbf{p}.\mathbf{z})}{ |\mathbf{v}|^2} \\
         \mathbf{p} = \frac{\mathbf{X_{goal}-X_{start}}}{|\mathbf {X_{goal}-X_{start}}|}, \quad  \mathbf{z} = \frac{\mathbf{z}}{|\mathbf{z}|}, \quad \mathbf{v} = \mathbf{p} \times \mathbf{z} \\
             [v]_x = \begin{bmatrix}
0 &  -v_z& v_y\\ 
 v_z&0  &-v_x \\ 
 -v_y& v_x & 0 
\end{bmatrix}
    \end{aligned}   
\end{equation} where $[v]_x$ is the skew-symmetric matrix of $\mathbf v$. Pose of ellipsoid ($X_{reduced}$) is represented in a compact form $\Sigma$ which allows to generate random samples within the $X_{reduced}$. In Algorithm~\ref{alg:random_samples_generate_reduced}, it is given how random samples are generated in which $\odot$ stands for element-wise multiplication. 

\begin{algorithm}
\caption{Generating samples from $X_{reduced}$}
\label{alg:random_samples_generate_reduced}
\begin{algorithmic}[1]
\Procedure{SampleFree}{}
    \State $covm \gets \Sigma$
    \State $npts \gets $ Number of random points (i.e, 500)
    \State $ndims \gets $Dimension size of the space (i.e, 3)
    \State $center \gets $ Center position of the ellipsoid
    \State $v,e \gets eig(covm)$
    
    \For {i=1,...,npts}
        \State $unit\_balls_i \gets \sim N(0, 1)$
        \For {j=1,...,ndims pt}
            \State $pt(i,j) \gets \sim N(0,1)$
        \EndFor
    \EndFor
    \State $pt_{opt} \gets (pt^t\odot 2)$
    \For {j=1,...,npts}
            \State $fac(j) \gets \Sigma_{i=0}^{ndims} pt_{opt}(j, i)$
    \EndFor
    \State $fac \gets (unit\_balls \odot (1/ndims)) \odot \frac{1}{sqrt(fac')}$
    \State $pnts \gets zeros(npts, ndims)$
    \State $d \gets \sqrt{(diag(e))}$
    \For {i=1,...,npts}
        \State $pnts(i,:) \gets fac(i)pt(i,:)$
        \State $pnts(i,:) \gets (pnts(i,:) \odot d'*v') + center $
    \EndFor
 \State \textbf{return} $pnts$
\EndProcedure
\end{algorithmic}
\end{algorithm}

As we stated, when forming the problem statement, we are interested in feasible path planning rather than optimal path planning.
This guarantees algorithm probabilistically complete but it might not be asymptotically optimal. If the planner probabilistically completes: $lim_{n \to \infty} \mathbb{P}({ V_{n} \cap X_{goal} \neq \varnothing; \sigma(0) = X_{start}, \sigma(1) \in cl(X_{goal})})$ should be equal to one. 
To achieve real-time performance, the planner should be capable of generating a feasible path within milliseconds, in that asymptotic optimality should be guaranteed. 
Asymptotic optimality is defined as each of the positions (waypoints) in path $\sigma(\tau)$ has clearance $\zeta$ to its closest obstacle. Hence, we further improved RRT* planner by proposing a technique to satisfy the asymptotical optimality, complemented by having real-time constraints.
Proposed method is given in Algorithm.~\ref{alg:asym}, that always checks the clearance ($\zeta$) when choosing $x_{nearest}$ point while incorporating the trajectory as shown in Fig.~\ref{fig:closest_point}. 
$\zeta$ depicts the distance between $x_{nearest}$ and closed waypoint on the trajectory. This is the second improvement that has been introduced to original RRT* (Algorithm~\ref{alg:rrt}: line 6). An example of path generations from original RRT* and improved RRT* is shown in Fig.~\ref{fig:result_of_rrt_and_improved}.
 
\begin{algorithm}
\caption{Selecting proper value for $x_{nearest}$ while satisfying asymptotic optimality}
\label{alg:asym}
\begin{algorithmic}[1]
\Procedure{Nearest}{$G=(V,E), x_{rand}, Traj$}
    \State $nearest_{obs} \gets ClosestObstacle(x_{rand})$
    \State $nearest_{waypoint} \gets ClosestWayPoint(nearest_{obs})$
    \State $center \gets nearest_{waypoint}$
    \State $npts \gets 10$
    \State $max\_attempt \gets 2$
    \State $radius \gets 4.0$
    \State $attempt \gets 0$
    \While {$i <= max\_attempt$}
        \State $covm(0.0)=covm(1,1)=covm(2,2) \gets radius$
        \State $random\_points \gets SampleFree()$
        \For {j=1,...,npts}
            \State $dis \gets |random\_points_{i}-nearest_{obs}$
            \If{$dis > obstacle\_fail\_safe\_dis$}
                \If{$CollisionFree(random\_points_{i})$}
                    \State \textbf{return} $random\_points_{i}$
                \EndIf
            \EndIf
        \EndFor
        \State $attempt \gets attempt+1$
        \State $radius \gets radius*2$
    \EndWhile
    
 \State \textbf{return} $pnts$
\EndProcedure
\end{algorithmic}
\end{algorithm}

\begin{figure}[ht]
\begin{center}
\includegraphics[width=5cm]{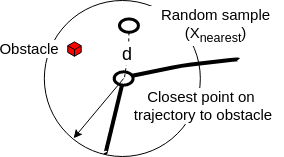}
\caption{Selecting proper value for $x_{nearest}$ while considering closest point to the trajectory from obstacle where $d$ denotes the clearance $\zeta$ which is a configurable parameter. Selecting $x_{nearest}$ is given in Algorithm.~\ref{alg:asym}}
\label{fig:closest_point}
\end{center}
\end{figure}

\begin{figure}[th!]
\begin{center}
\includegraphics[width=\linewidth]{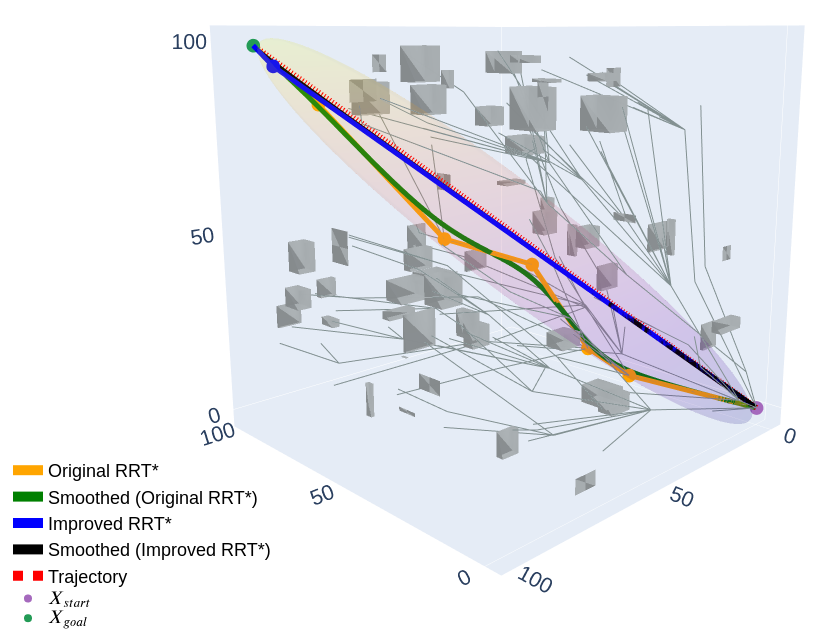}
\caption{Path generated by original RRT* and corresponding B-spline trajectory is depicted in orange and green colors respectively, whereas paths that are denoted in blue color and back color correspond to path and smoothed path of improved RRT*}
\label{fig:result_of_rrt_and_improved}
\end{center}
\end{figure}

\section{Experimental Results}

In this section we present a complete validation of the proposed long range local replanner. First, we validate our proposed instance map building with two other techniques: octomap~\cite{hornung2013octomap} which is one of the most popular techniques among the researches and 3D Circular Ring Buffer~\cite{usenko2017real}. 
Secondly, performance of improved RRT*, original RRT*(~\cite{karaman2011sampling}) and A* are compared. Complete system is implemented in C++11 in that the sparse matrix library Eigen is used while enabling GNU C++ compiler optimization level to -O2. Finally, complete system is evaluated in the simulated and the real environments. The simulation is done on a laptop equipped with a 2.20 GHz Intel Core i5-5200U CPU and 8 GB RAM. The flight experiments are done on a DJI M600 quadrotor( Fig.~\ref{f:main_steps}a) equipped with an Intel NUC (dual-core CPU i5-4250U at 1.30 GHz and 16GB RAM). 

\subsection{Instance Map Building}
In our proposed approach, initial step is to filter out regions that interest from each Lidar point cloud. Afterwards, consecutive regions of interest are merged. 
First step of trajectory estimation thread is to extract regions with the $obs\_avoid\_dis$ from current pose of MAV and push into a Rtree. That is how instance map is constructed. We have used 10000 lidar point clouds and calculated average time of insertion. Result is shown in Fig.~\ref{fig:comparison_of_inserstion}. According to result, execution time for our propose solution is much less than for the OctoMap, whereas 3D ring buffer performs slightly better than ours.


\begin{figure}
\subfloat[OctoMap~\cite{hornung2013octomap}]{  \includegraphics[width=\linewidth, height=3cm]{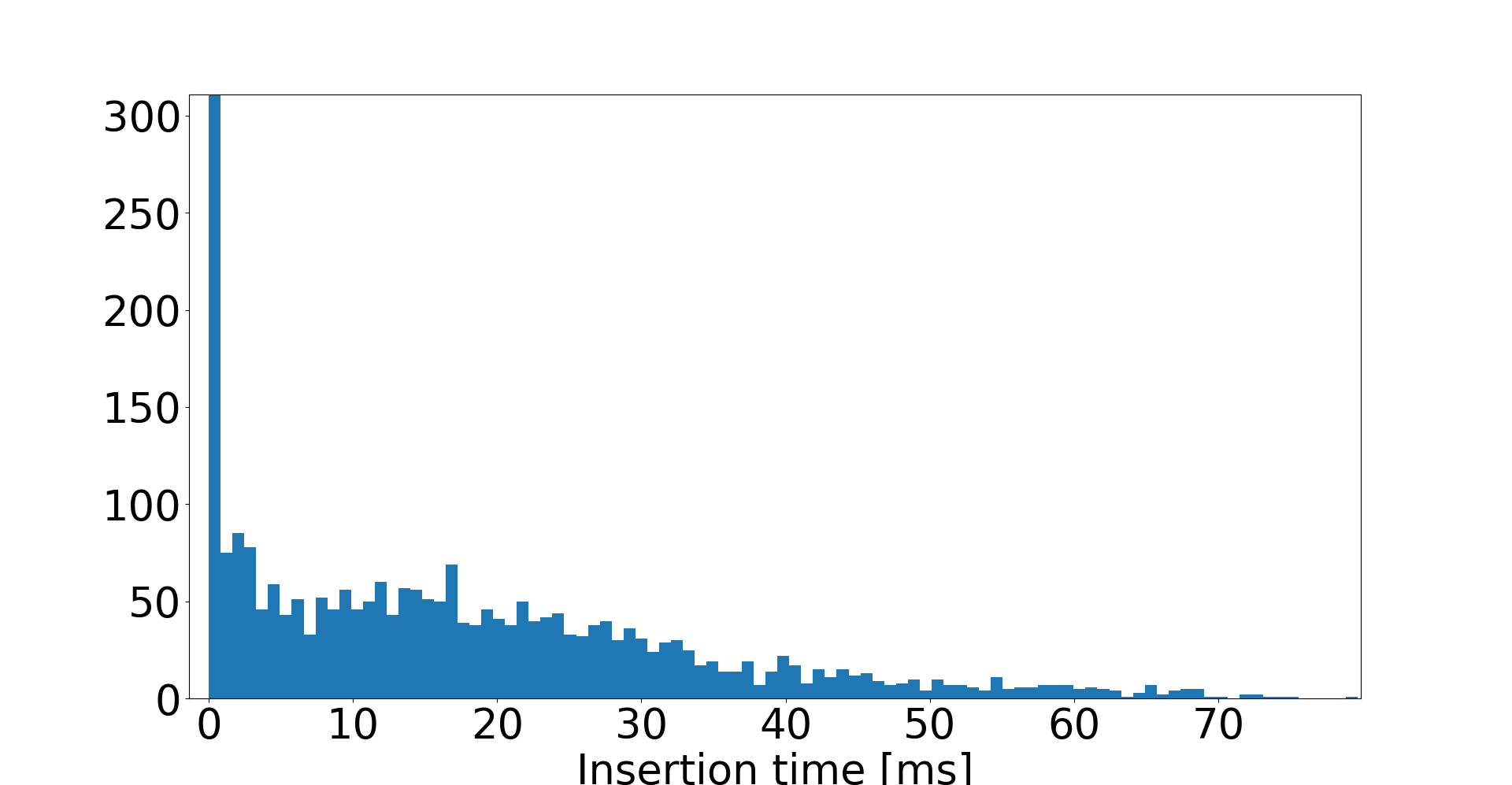} 
\label{fig:ocatomap}}

\subfloat[3D Ring Buffer~\cite{usenko2017real}]{  \includegraphics[width=\linewidth, height=3cm]{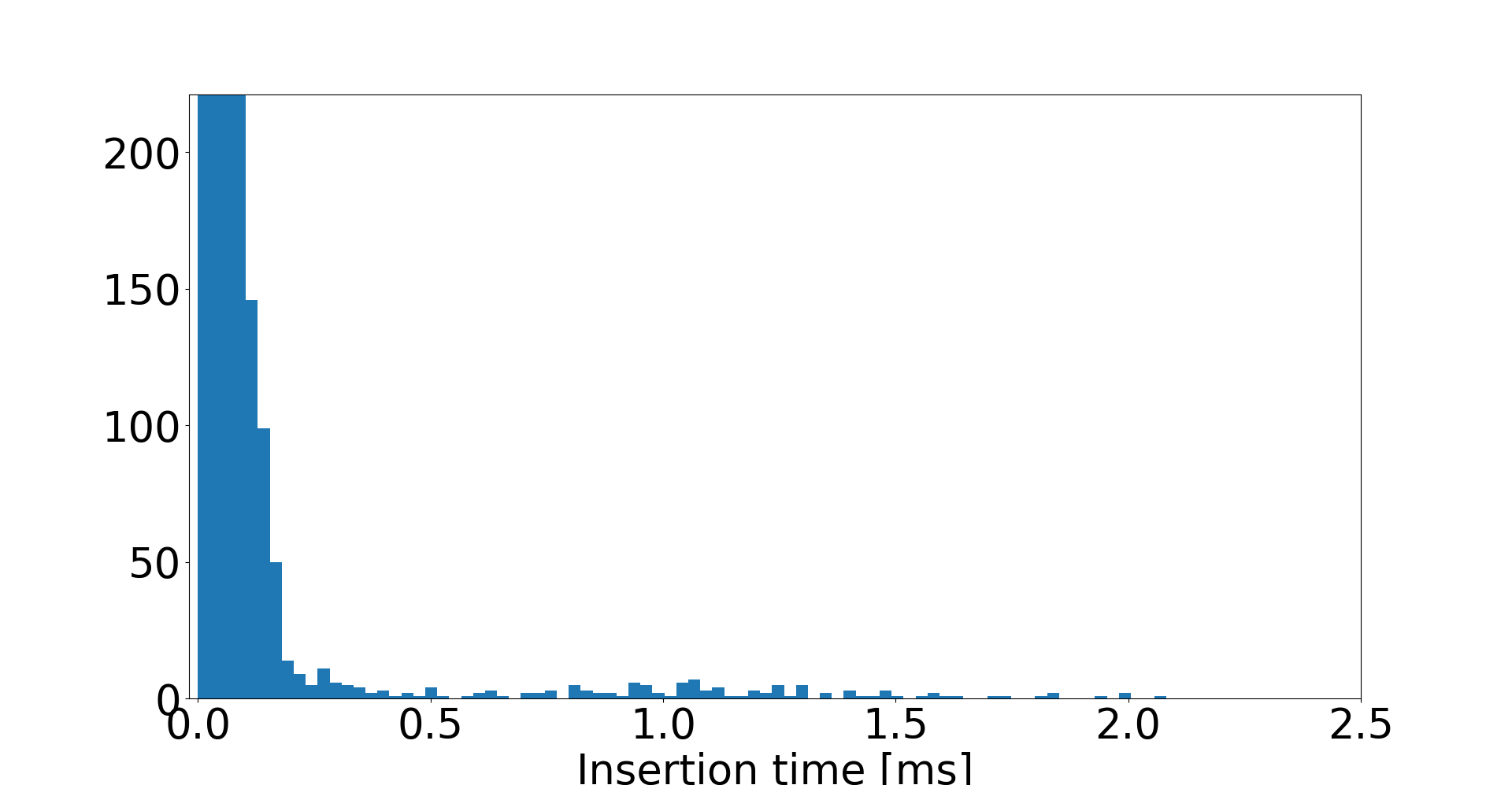} \label{fig:ringbuffer}}

\subfloat[Proposed instance map building approach]{  \includegraphics[width=\linewidth, height=3cm]{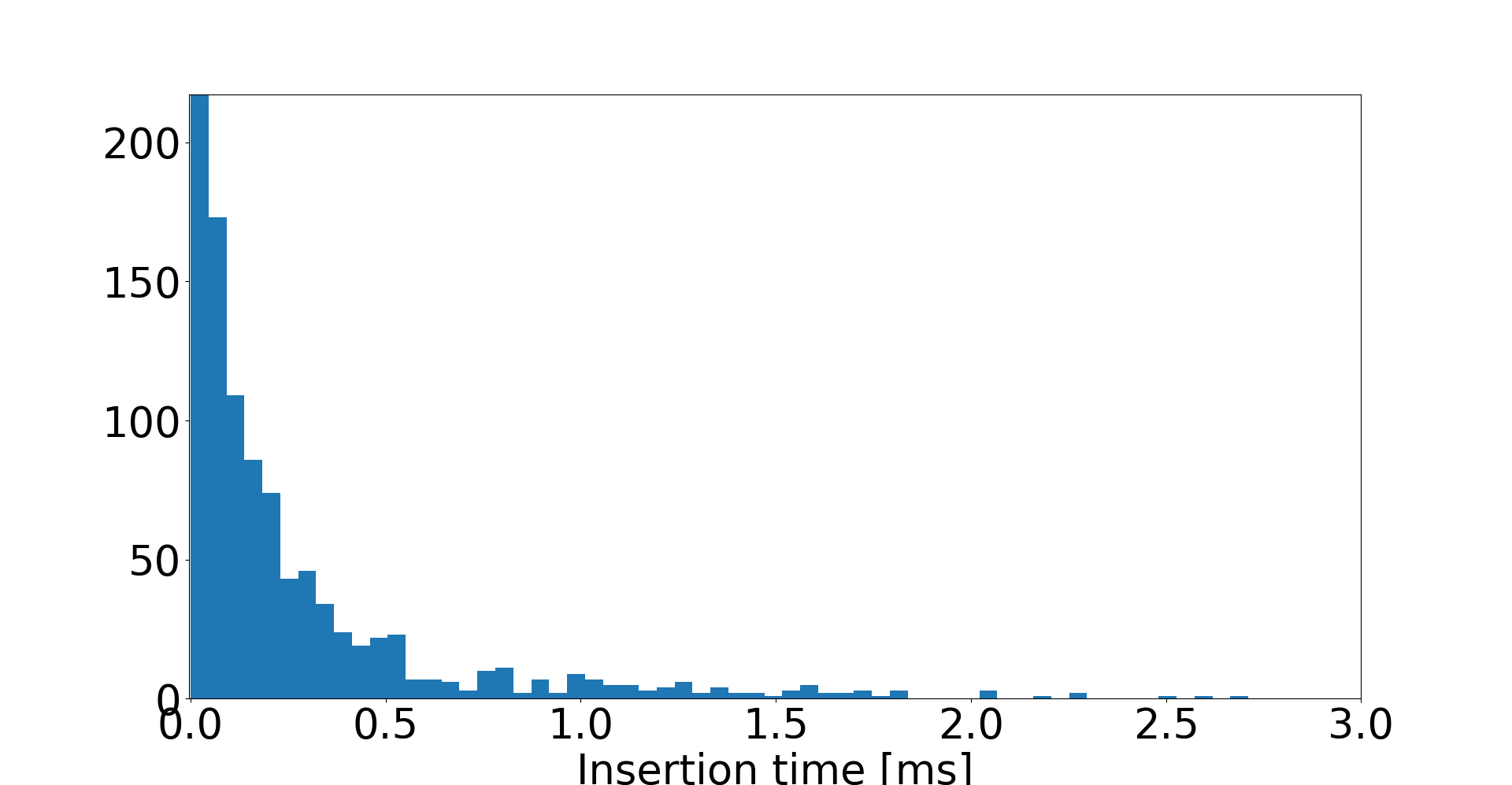} \label{fig:ourapproach}}

  \caption{Comparison of histograms of insertion time for occupancy mapping. Obstacle avoidance zone (Lidar range) is taken as 5m in which resolution of the map is set to 20cm}
  \label{fig:comparison_of_inserstion}
\end{figure}

\subsection{3D Planner Performance}
For evaluate performance of the proposed planner, we developed 3D A* planner in addition to RRT* and improved RRT*.
All three algorithms utilize the same search space. 
For performance analyzes, we ran considered algorithms 100 times for a given search space. Likewise, we tested 10 different search spaces in which dimension on each direction (x,y and z) kept at 10m. Also, start and goal positions are kept at same positions. Results are given in Table.~\ref{t:time_estimation}. 
In all the cases, improved RRT* takes minimum amount of time for finding a path. Usually RRT* is used for global planning. After the modifications we made, it is capable of finding path within a few milliseconds which lead us to employ this for local replanning. To calculate cost of the smoothed path, we define a different kind of norm $U_{path}$(\cite{opt}). And its gradient can be derived as given in Eqs.~\ref{eq:cost_calculation1} and in Eqs.~\ref{eq:cost_calculation} in matrix form, assuming, path consists of n number of points.  

\begin{equation}
    \begin{aligned}
         U_{path} = \frac{1}{2} \Sigma_{i=0}^n|q_{i+1}-q_i| \\
         \frac{\partial U_{path}}{\partial q_i} = 2q_i -q_{i+1}-q_{i-1}
    \end{aligned}
    \label{eq:cost_calculation1}
\end{equation}
\begin{equation}
   \frac{\partial U_{path} }{\partial q_i}  =  Aq = \begin{bmatrix}
 2 & -1&  0&  0& ...\\ 
 -1&  2&  -1&  0& ...\\ 
 0&  -1&  2&  -1& ...\\ 
 \vdots &  \vdots &  \vdots & \vdots & \ddots \\ 
 0&  0&  0&  -1& 2
\end{bmatrix}\begin{bmatrix}
q_1\\ 
q_2\\ 
q_3 \\ 
\vdots \\ 
q_n
\end{bmatrix}
\label{eq:cost_calculation}
\end{equation}

Thus, cost of each path can be found by $q^tAq$. Since, desired trajectory($T_{online}$) is known, we define the path cost as follows:
\begin{equation}\label{eq:path_cost}
   path_{cost} = \frac{q^tAq}{T_{online}^tAT_{online}}
\end{equation} Results are given in Table~\ref{t:time_estimation}.

\begin{figure}[ht]
\begin{center}
  \includegraphics[width=\linewidth, height=6cm]{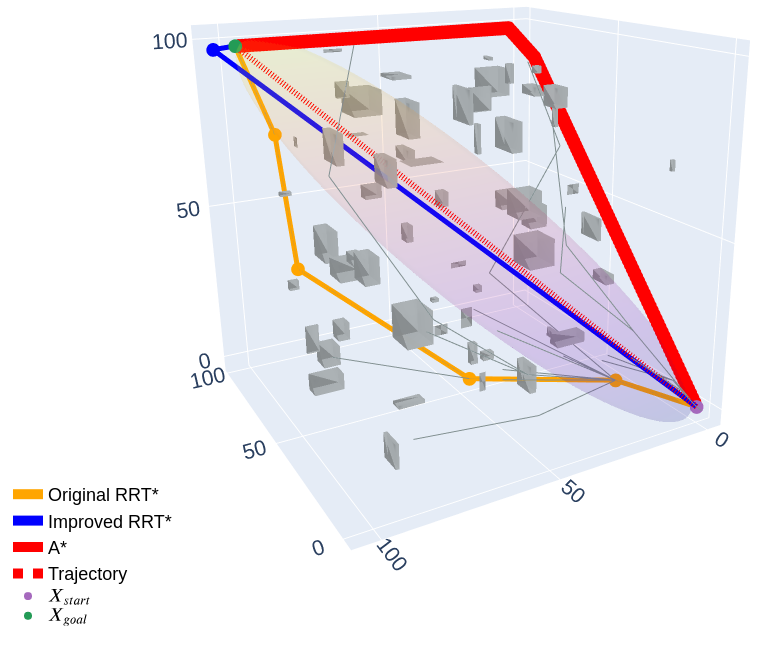}
\end{center}
\caption{Improved RRT* finds a path very closer to the given trajectory(red dashed line). A* takes less time, cost of the path much higher than even RRT*. This result corresponds to the 3rd test result given in Table.~\ref{t:time_estimation}}
\label{fig:moreobs}
\end{figure}

\begin{table}[ht]
  \caption{Search space set as 10m on each direction (x,y and z) with fix a set of obstacles (Fig.~\ref{fig:moreobs}) in that voxel size is set to 0.1m. Afterwards, improved RRT* (proposed), original RRT* and A* algorithms were executed 10 times. Execution times and path costs are shown}\label{t:time_estimation}
  \begin{center}
    \begin{tabular}{|c|c|c|c|c|c|c|}
    \hline
    \multirow{2}{*}{No.}& \multicolumn{3}{|c|}{Time (ms)} & \multicolumn{3}{c|}{$path_{cost}$} \\
      & A*  & RRT* & Ppsd  & A*  & RRT* & Ppsd \\
    \hline
  1& 12.45&180.91&1.56  & 1.46&1.32&0.78 \\
\hline
  2&9.45&149.96&1.39 & 1.23&1.29&0.79\\
\hline
 3&17.99&558.88&3.75 & 1.58&1.34&0.79\\
\hline
4&9.76&103.60&1.65  & 1.78&1.27&1.00\\
\hline
5&10.12&557.22&1.27& 1.34&1.35&0.82\\
\hline
6&10.13&161.70&1.74  & 1.260&1.27&0.81\\
\hline
7&9.68&548.10&1.26 & 1.89&1.31&0.78\\
\hline
8& 10.49&98.14&1.37 & 1.45&1.28&0.80\\
\hline
9 &9.84&2300.89&1.60 & 1.28&1.28&0.79\\
\hline
10 &9.47&111.61&1.63 & 1.89&1.31&0.79\\
\hline
  \end{tabular}
  
  \end{center}
\end{table}

\subsection{Evaluation of long range trajectory follower}
Firstly, MAV is able to fly autonomously given any feasible trajectory. Thus, We have evaluated our trajectory in a obstacle free zone. Replanning is not required in obstacle free zone. Thus, MAV should fly on given trajectory. To estimate error of trajectory following, MAV was tested on 5 different trajectories. The cost of the path was calculated using Eqs.\ref{eq:path_cost}. Average value of cost is 0.45 $\pm$ 0.14. One of the testing trajectories and corresponding MAV trajectory is shown in Fig.\ref{fig:labeing_trajectory}.

\begin{figure}[ht]
\begin{center}
\subfloat[One of the testing trajectories MAV should follow]{ \includegraphics[width=0.95\linewidth, height=1.35cm]{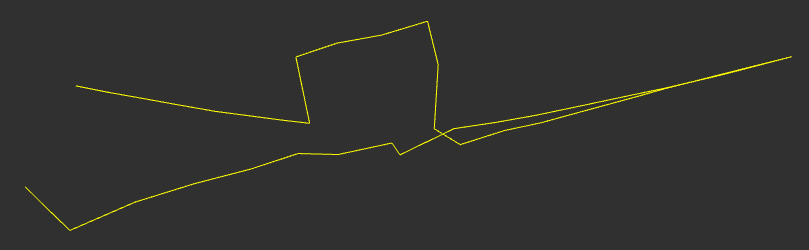}
\label{f:trajectory}} \\
\subfloat[The trajectory which MAV navigates autonomously is shown in purple colour small ellipsoids where no obstacles are present]{  \includegraphics[width=0.95\linewidth, height=1.35cm]{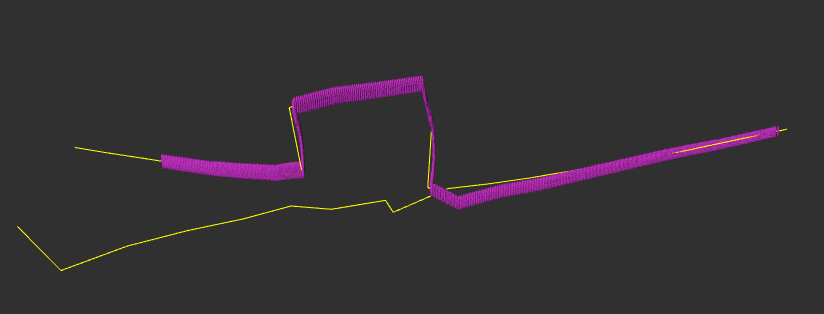} \label{f:path_of_mav}}
  \end{center}
  \caption{Example scenario for evaluation of trajectory follower}
 \label{fig:labeing_trajectory}
\end{figure}

\subsection{Testing the complete system}
We have tested our complete system in simulated and real environments. Simulator is developed based on Unity 3d which provides simulating the flying environment, complemented with 16 channels Lidar which gives the MAV surrounding as a point cloud. Two examples are shown in Fig.~\ref{fig:test_result_2}. For real word testing, we used DJI M600 hexacopter with Velodyne VLP16 Lidar, shown in Fig.~\ref{f:separation_on_bfs}. Behaviour of drone when it detects dynamic obstacle were experimented.

\begin{figure}[ht]
\begin{center}
\subfloat[]{ \includegraphics[width=0.95\linewidth, height=2.7cm]{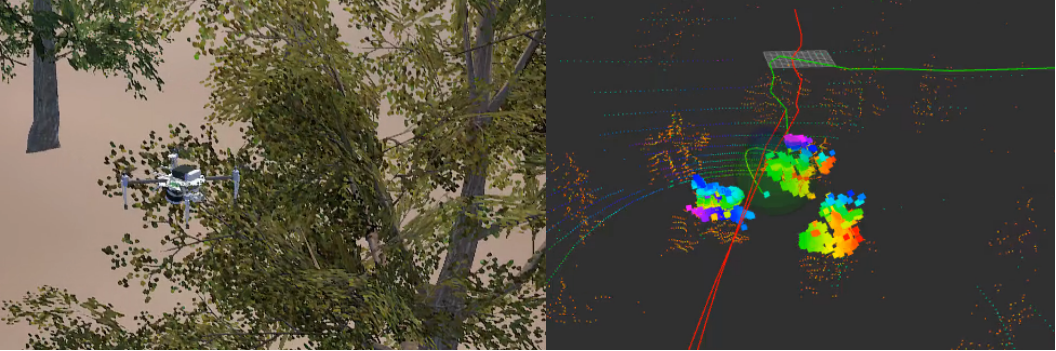}
\label{f:test_result_1}} \\
\subfloat[]{ \includegraphics[width=0.95\linewidth, height=2.7cm]{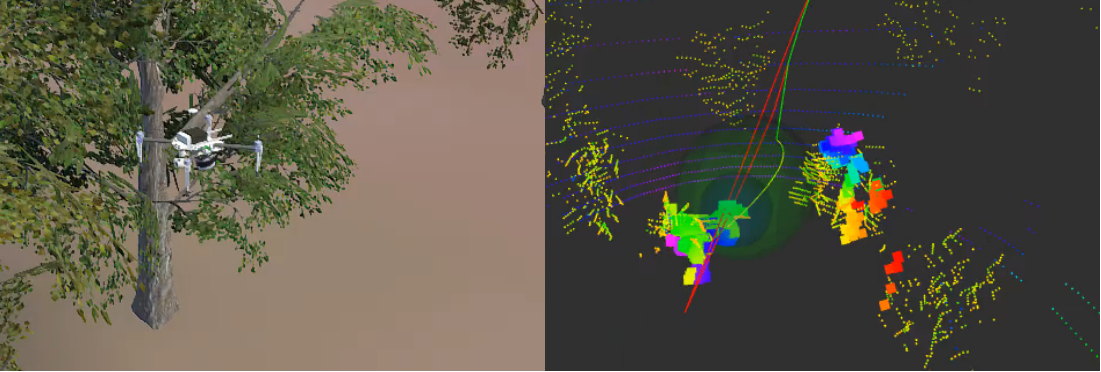}}
  \end{center}
  \caption{Sample scenarios of replanning the trajectory in a cluttered environment}
 \label{fig:test_result_2}
\end{figure}

\section{Conclusion}
We have developed and tested the complete system for real-time long-range local replanning presence of dynamic obstacles in real and simulated environments. Proposed improvements for original RRT* helped us to achieve real-time performance. Still, there is room for improvement which we noticed while testing. We are going to improve the proposed planner by incorporating kinodynamic motion planning constraints to increase the confident level of the planner.  

\section*{Acknowledgment}
The work presented in the paper has been supported by Innopolis University, Ministry of Education and National Technological Initiative in the frame of creation Center for Technologies in Robotics and Mechatronics Components (ISC 0000000007518P240002)

\bibliographystyle{IEEEtran}
\bibliography{ifacconf}

\end{document}